\documentclass{article}

\usepackage[numbers]{natbib}

\usepackage[final]{neurips_2024}

\usepackage[utf8]{inputenc} 
\usepackage[T1]{fontenc}    
\usepackage{hyperref}       
\usepackage{url}            
\usepackage{booktabs}       
\usepackage{amsfonts}       
\usepackage{nicefrac}       
\usepackage{microtype}      
\usepackage{xcolor}         
\usepackage{soul}
\usepackage{booktabs}
\usepackage{graphicx}
\usepackage{caption}
\usepackage{subfigure}
\usepackage{tabularx}
\usepackage{makecell}
\usepackage{algorithm}
\usepackage{algorithmic}
\usepackage{bbm}
\usepackage{amsmath}
\usepackage{amssymb}
\usepackage{mathtools}
\usepackage{amsthm}
\usepackage{colortbl}
\usepackage{wrapfig}
\usepackage[
  separate-uncertainty = true,
  multi-part-units = repeat
]{siunitx}

\DeclareMathOperator*{\argmin}{arg\,min}
\DeclareMathOperator{\expect}{\mathbb{E}}

\title{Policy Improvement using Language Feedback Models}

\author{%
  Victor Zhong\thanks{Corresponding author.} \\
  University of Waterloo\\
  Microsoft Research\\
  \texttt{victor.zhong@uwaterloo.ca} \\
  \And
  Dipendra Misra\\
  Microsoft Research\\
  \And
  Xingdi Yuan\\
  Microsoft Research\\
  \And
  Marc-Alexandre C\^{o}t\'{e}\\
  Microsoft Research\\
}

\begin{document}
\newcommand{\mathsc}[1]{{\normalfont\textsc{#1}}}

\newcommand{\titleshort}{Policy Improvement using Language Feedback Models}
\newcommand{\titlelong}{\titleshort}
\newcommand{\methodname}{Language Feedback Model}
\newcommand{\methodnameshort}{LFM}

\newcommand{\instruction}{x}
\newcommand{\state}{s}
\newcommand{\observation}{o}
\newcommand{\verbalized}{v}
\newcommand{\verbalize}{V}
\newcommand{\action}{a}
\newcommand{\expertaction}{\hat{\action}}
\newcommand{\traj}{\tau}
\newcommand{\policy}{\pi}
\newcommand{\expert}{\policy^*}
\newcommand{\param}{\theta}

\newcommand{\env}{E}
\newcommand{\rollout}{{\mathsc{Rollout}}}
\newcommand{\dataset}{D}
\newcommand{\feedbackdataset}{F}
\newcommand{\feedbackmodel}{f}
\newcommand{\feedback}{y}
\newcommand{\feedbacklabel}{y^*}
\newcommand{\llmmethod}{{\mathsc{Llm}}}
\newcommand{\desirable}{{\mathsc{Desirable}}}
\newcommand{\loss}{{L}}

\newcommand{\true}{{\mathrm{yes}}}
\newcommand{\false}{{\mathrm{no}}}
\newcommand{\indicator}{{\mathbbm{1}}}

\newcommand{\imitationdataset}{G}

\newcommand{\methodBehaviourCloning}{{\textsc{Bc}}}
\newcommand{\methodActionPrediction}{{\textsc{ActPred}}}
\newcommand{\methodOurs}{{\textsc{Lfm}}}
\newcommand{\methodOursAdapt}{{\textsc{LfmA}}}
\newcommand{\methodOursDetailed}{{\textsc{LfmD}}}

\definecolor{darkgreen}{HTML}{548235}
\definecolor{darkred}{HTML}{C00000}
\definecolor{darkblue}{HTML}{2e75B6}
\definecolor{darkyellow}{HTML}{BF9000}
\definecolor{darkpurple}{HTML}{7030A0}
\definecolor{lightgray}{HTML}{f3f3f3}
\definecolor{lightergray}{HTML}{fbfbfb}
\definecolor{lightergreen}{HTML}{f9fff9}
\definecolor{lighteryellow}{HTML}{fffff9}

\newcommand{\todo}[1]{\textcolor{red}{TODO: #1}}

\maketitle

\begin{abstract}
We introduce~\methodname s (\methodnameshort s) that identify desirable behaviour --- actions that help achieve tasks specified in the instruction --- for imitation learning in instruction following.
To train~\methodnameshort s, we obtain feedback from Large Language Models (LLMs) on visual trajectories verbalized to language descriptions.
First, by using~\methodnameshort s~to identify desirable behaviour to imitate, we improve in task-completion rate over strong behavioural cloning baselines on three distinct language grounding environments (Touchdown, ScienceWorld, and ALFWorld).
Second, imitation learning using~\methodnameshort s~outperform using LLMs as experts to directly predict actions, when controlling for the number of LLM output tokens.
Third,~\methodnameshort s~generalize to unseen environments, improving task-completion rate by 3.5-12.0\% through one round of adaptation.
Finally, we modify~\methodnameshort s to provide human-interpretable feedback without performance loss, allowing human verification of desirable behaviour for imitation learning.
\end{abstract}

\section{Introduction}

\begin{figure*}[!t]
\centering     
\subfigure[
Learning a small and cost-effective~\methodname~from LLM feedback.
We roll out an initial policy, then prompt an LLM to provide feedback on what actions the policy took during the rollout were productive in achieving the task outlined in the instruction.
We then use this data to train a feedback model that predicts whether an action is productive given the instruction.
]{
    \label{fig:feedback_model}
    \includegraphics[width=\linewidth]{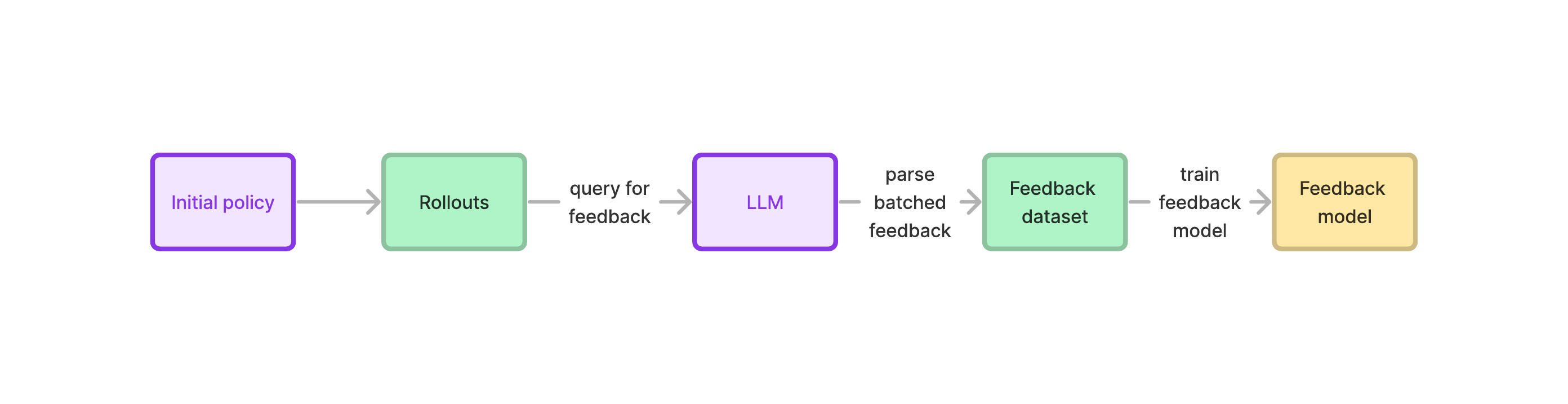}
}

\subfigure[
Policy improvement by imitating desirable behaviour identified by a learned feedback model.
Given the instruction, we roll out a base policy, then identify productive actions that help achieve tasks specified in the instruction using the trained feedback model.
Finally, we update the base policy by imitating productive actions.
]
{
    \label{fig:policy_improvement}\includegraphics[width=\linewidth]{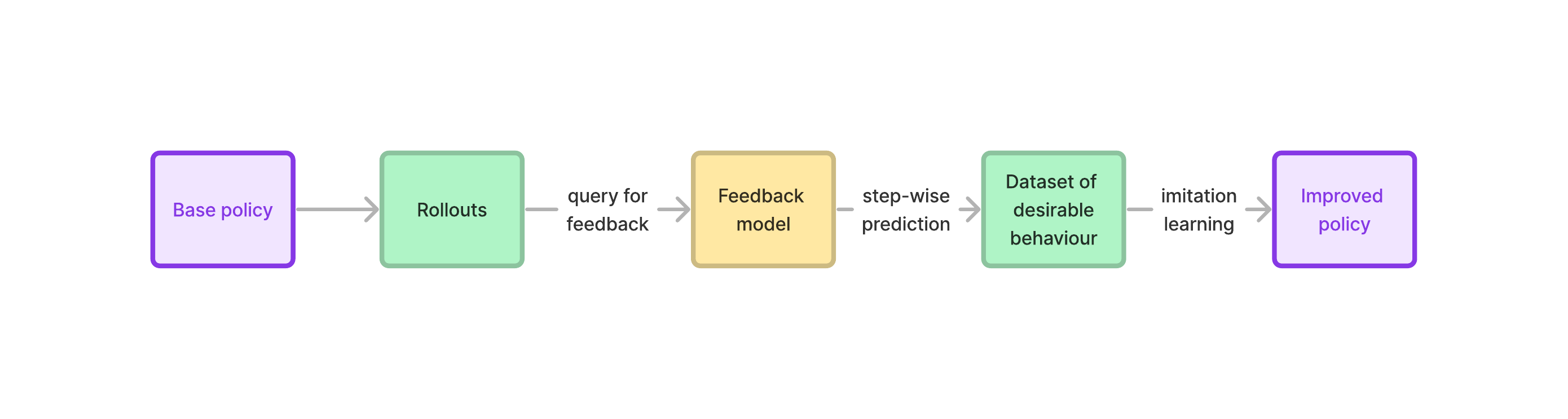}
}
\subfigure[
Example of desirable behaviour identified in ALFWorld, a kitchen instruction following benchmark.
]
{
    \label{fig:comic}\includegraphics[width=\linewidth]{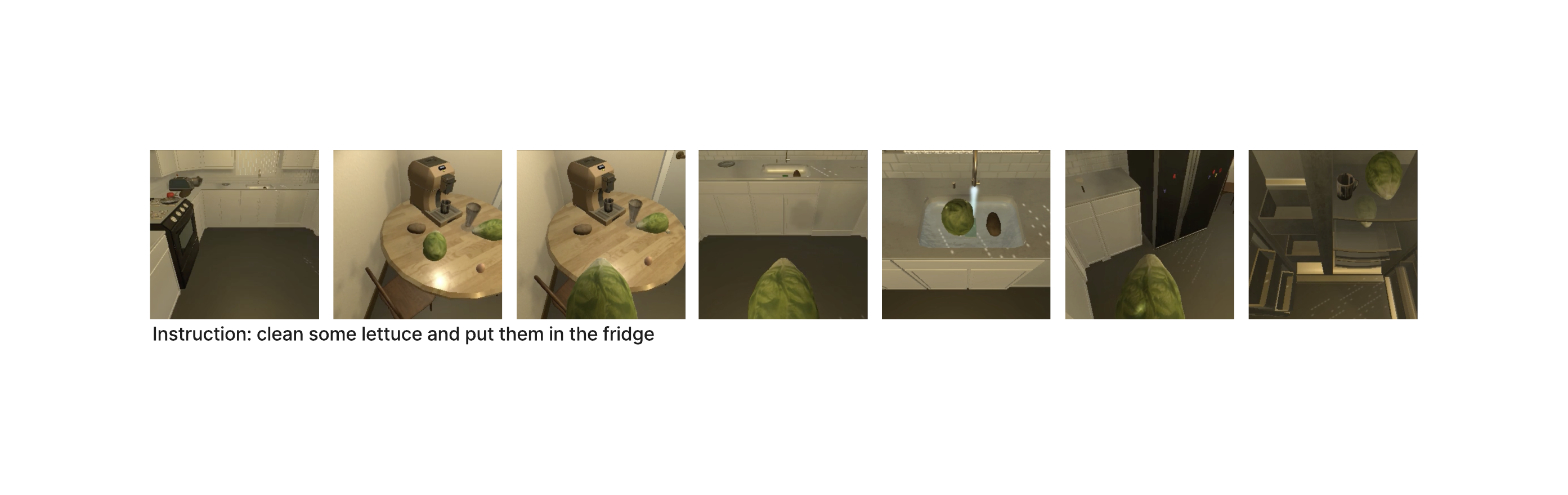}
}
\vspace{-0.1in}
\caption{
Given an environment and instructions to follow, we assume a verbalization procedure that converts observations to language descriptions.
Policy improvement using~\methodname~involves (a) training a feedback model, then (b) using it to identify desirable behaviour for policy improvement via imitation learning.
The feedback model is \textcolor{darkyellow}{yellow}, other models \textcolor{darkpurple}{purple}, and generated intermediate data~\textcolor{darkgreen}{green}.
An example of~\methodnameshort-identified behaviour is shown in (c).
}
\label{fig:main}
\vspace{-0.2in}
\end{figure*}

Sample-efficiency and generalizability are two primary challenges in learning instruction following agents in grounded environments~\citep{macmahonWalkTalkConnecting2006,kollarUnderstandingNaturalLanguage2010,ahnCanNotSay2022}.
First, we want an agent that is sample-efficient: it learns from few demonstrations of how to act according to instructions.
Second, we want an agent that is generalizable: it should act successfully in novel environments according to new instructions after training.
Reinforcement learning (RL;~\citet{Sutton1998}) and imitation learning (IL;~\citet{schaal1999IsImitationLearning},~\citet{abbeel04ApprenticeshipLearning}) are two techniques for learning agents for instruction following in grounded environments.
These techniques often require large numbers of trials and errors or expensive-to-obtain expert demonstrations.
Recent work show that pretrained large language models (LLMs) exhibit sample-efficient learning through prompting and in-context learning for textual~\citep{brownLanguageModelsAre2020} and grounded problems such as robotic control~\citep{ahnCanNotSay2022}.
However, for instruction following in grounded problems, current methods rely on LLMs on-line during inference, which is impractical and expensive.

We develop a sample-efficient and cost-effective technique that uses LLMs to train~\textbf{\methodname s} (\textbf{\methodnameshort s}) for policy improvement in instruction following.
Figure~\ref{fig:main}~illustrates policy improvement using~\methodnameshort s.
Consider the task of interacting with objects in a kitchen to follow instructions shown in Figure~\ref{fig:comic}.
First, in Figure~\ref{fig:feedback_model},
given a grounded environment and a base policy (i.e.~a behaviour cloned policy), we roll out the base policy to collect a small set of trajectories for different instructions.
Next, we verbalize observations in the trajectory by describing scenes in language.
For each instruction and verbalized trajectory pair, we query an LLM to provide feedback identifying which behaviour in the trajectory is productive to solving the task identified in the instruction (i.e.~answer yes or no).
For instance, given an instruction ``put a clean slice of lettuce in the refridgerator'',
GPT-4~\citep{openaiGPT4TechnicalReport2023}~is able to deduce that key milestones are 1) find the lettuce, 2) slice it 3) wash it in the sink, and 4) put it in the fridge.
Consequently, such an LLM is able to identify when an agent is exhibiting~\textbf{desirable behaviour} conducive to solving tasks outlined in the instruction, for instance by taking the lettuce to the sink, versus undesirable behaviour, for instance by cooking the lettuce.
We define desirable behaviour as productive actions that are constructive, task-beneficial, and effective in following the instruction.
In other words, taking the action brings the agent closer (in terms of trajectory length) to accomplishing the task specified in the instruction.
After collecting LLM feedback, we distill this world knowledge into a small and cost-effective~\methodnameshort.
Finally, in Figure~\ref{fig:policy_improvement}, given a policy to improve on potentially new environments and instructions, we use the learned~\methodnameshort~to identify desirable actions on-line, then update the policy to imitate these actions.
Crucially, this technique is sample-efficient in that it improves policy with no additional human-labeled demonstrations.
Furthermore, this technique is cost-effective in that it requires few LLM interactions to collect an off-line dataset during~\methodnameshort~training (i.e.~before deployment), as opposed to many LLM interactions on-line during policy improvement (i.e.~after deployment).

Our findings are as follows:
first,~\methodnameshort~policy improvement achieves consistent gains over strong behaviour cloned base policies on three grounded instruction following benchmarks in Touchdown~\citep{chenTouchdownNaturalLanguage2019}, ScienceWorld~\cite{Wang2022ScienceWorld}, and ALFWorld~\cite{ALFWorld20}.
Second, we compare~\methodnameshort s~against prompting LLMs to directly predict what actions to take, then imitating this LLM-predicted behaviour.
On all benchmarks, using~\methodnameshort~feedback outperforms using LLMs as experts for imitation learning, given a fixed allocation of LLM output tokens.
This gain is especially pronounced in environments with larger action spaces, such as ScienceWorld, where it is much easier to critique than to generate the correct action.
Third, we show that learned feedback models generalize to unseen environments with new tasks and new transition functions.
After training~\methodnameshort s~on training environments, we use them to identify desirable behaviour on test environments, which we then imitate to adapt the policy.
A single round of adaptation achieves significant gains (3.5-12.0\% task-completion) across all environments.

In addition to policy improvement, using~\methodnameshort~feedback offers two advantages over existing techniques such as using LLMs as expert policies for imitation learning.
First,~\methodnameshort~improves policies on-line without additional expensive calls to LLMs.
Second,~\methodnameshort~can offer human-interpretable feedback when identifying desirable behaviour to imitate.
We show in Section~\ref{sec:results} that~\methodnameshort s~can be easily modified to provide not only desirable behaviour but why they were desirable, thereby allowing humans to inspect and validate imitation data used for policy improvement.
Source code for our environments and experiments are available at~\href{https://github.com/vzhong/language_feedback_models}{github.com/vzhong/language\_feedback\_models}.
Videos of~\methodnameshort~feedback are available at~\href{https://language-feedback-models.github.io/}{language-feedback-models.github.io}.

\begin{figure*}[!t]
    \includegraphics[width=\linewidth]{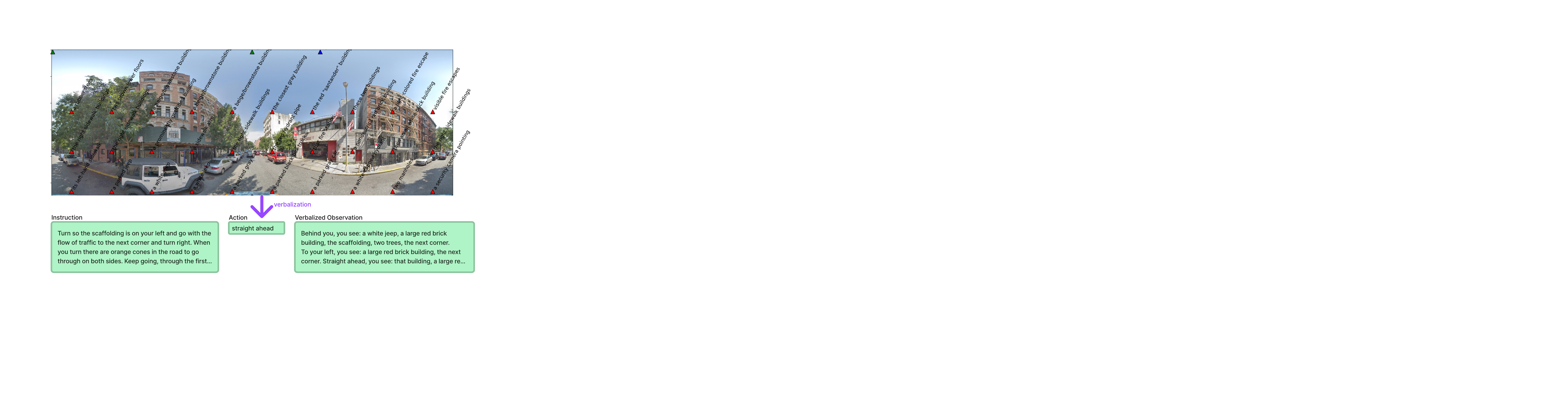}
    \vspace{-0.15in}
    \caption{
    An example verbalization for Touchdown.
    We align~\mathsc{Clip} image embeddings of panorama patches and language embeddings of common noun-phrases to populate a language template.
    Appendix~\ref{app:touchdown_verbalization}~describes this procedure in detail.
    The~\textcolor{darkblue}{blue} arrow at the top indicate the agent's orientation while the~\textcolor{darkgreen}{green} arrows indicate valid directions to proceed in.
}
    \label{fig:touchdown_verbalization}
\vspace{-0.1in}
\end{figure*}

\section{Background}
\paragraph{Language grounded instruction following.}
In language-grounded instruction following, an agent is given an instruction $\instruction$ specifying the task to achieve in the environment.
Each turn, the agent receives a potentially partial observation $\observation_t$, and takes an action $\action_t$ which causes the environment to transition to a new state.
In the Figure~\ref{fig:comic} example, the agent observes a counter with objects such as a toaster, some lettuce, and a knife on top.
To follow the instruction ``put a clean slice of lettuce in the refridgerator'', an effective agent may choose to grab a piece of lettuce.
In the reinforcement learning setting, the environment additionally give the agent a reward after a desirable (positive reward) or undesirable (negative reward) action~\citep{Sutton1998}.
In this work, we consider long-horizon settings with only sparse and delayed task-completion rewards.
Consequently, we focus on imitation learning from demonstrations as opposed to reinforcement learning from rewards~\citep{schaal1999IsImitationLearning}.

\paragraph{Online imitation learning.}
In online imitation learning for instruction following, we are given an expert policy $\expert(\action|\instruction, \observation)$ and learn a policy $\policy_\param(\action|\instruction, \observation)$ with parameters $\param$.
We first roll out the policy $\policy_\param$.
For each step $\observation^{(i)}_t$ of the rollout $\traj_i$, we optimize $\param$ to imitate the action $\action^{(i)}_t$ chosen by the expert $\expert (\action|\instruction, \observation^{(i)}_t)$ when given the same observations:
$
    \argmin_\theta \expect_{\expert} \left[
        \loss \left( \policy_\theta(a|\instruction, \observation^{(i)}_{t}), \action^{(i)}_t \right)
    \right]
$.
Here, $\loss$ is step-wise cross-entropy between the policy's action distribution and the action chosen by the expert given the same observation:
%
$
    \loss \left( * \right)
    =
    -\sum_{\action^\prime \in \mathcal{A}} \indicator \left[\action^\prime = \action^{(i)}_t\right] \ln \policy_\theta(\action = \action^\prime \mid \instruction, \observation^{(i)}_t).
$
%
\paragraph{Behavioural cloning.}
Online imitation learning assumes an expert policy that can be executed online to produce expert actions.
For instance, given an expert, imitation learning assumes that this expert $\expert (\action|\instruction,\observation_t)$ provides corrective actions $a_t$ as the policy $\policy (\action|\instruction, \observation_t)$ runs.
In many cases, this is impractical --- a human-in-the-loop expert is expensive and inconvenient while an LLM expert is expensive and, as we show in our experiments, inaccurate.
Alternatively, in behaviour cloning (BC), we instead collect an offline dataset of expert trajectories from which to clone expert behaviour~\cite{Bain1995AFF,torabi2018behavioral}.
BC (or offline imitation learning) only asks the expert to perform the task $N$ times to collect $N$ trajectories $\{\traj_i\}_{i=1}^N$.
Each $\traj_i$ consists of $M_i$ steps of observations and associated expert actions: $\traj_i = [ \observation^{(i)}_1, \action^{(i)}_1,  \ldots, \observation^{(i)}_{M_i}, \action^{(i)}_{M_i}]$ where $\action^{(i)}_{t}$ is the action chosen by the expert $\expert (\action|\instruction, \observation^{(i)}_t)$ given the observation $\observation^{(i)}_t$.
We train policy $\policy_\theta$ to imitate the expert action, given the same observation seen by the expert, by minimizing the following objective:
%
$
    \argmin_\theta \frac{1}{N} \sum_i^N \frac{1}{M_i} \sum_t^{M_i} \loss \left( \policy_\theta (\action|\instruction, \observation^{(i)}_t), \action^{(i)}_t \right).
$
%
The key distinction between BC and imitation learning is that the former optimizes over trajectories under the expert policy while the latter optimizes over trajectories under the learned policy.
Consequently, while BC is offline and easily batchable, it suffers from covariate shift/exposure bias~\citep{rossReductionImitationLearning2011,bengioScheduledSamplingSequence2015}.
Like prior work in long-horizon instruction following in grounded environments~\citep{friedUnifiedPragmaticModels2018,chenTouchdownNaturalLanguage2019}, we use BC to warm-start a strong base policy~\citep{ashWarmStartingNeuralNetwork2020}, which we then improve using imitation learning.

\section{\methodname}

How can we leverage world knowledge in LLMs to make policy learning more sample-efficient and generalizable?
In this work, we use LLMs to distill a small and cost-effective~\methodname~to identify desirable behaviour from a base policy (Figure~\ref{fig:feedback_model}).
We then improve the base policy by imitating this desirable behaviour through batched imitation learning, without need for on-line LLMs (Figure~\ref{fig:policy_improvement}).
Appendix~\ref{app:pseudocode}~provides pseudo-code for the entire procedure for policy improvement using~\methodnameshort s.
A natural question is why not directly use LLMs as experts for action prediction.
Section~\ref{sec:results} shows that the using LLMs to learn feedback models results in higher policy improvement than using LLMs as experts for action prediction.
Moreover,~\methodnameshort s~generalize to new environments unseen during training, thereby allowing policy improvement on new environments.

\subsection{Verbalization}
\label{sec:verbalization}
To leverage world knowledge in LLMs, we convert raw observations $\observation$ to language descriptions $\verbalized$ using a verbalization procedure $\verbalize$.
Figure~\ref{fig:touchdown_verbalization}~illustrates verbalization for Touchdown~\cite{chenTouchdownNaturalLanguage2019}, where the agent navigates Google Street View panorama images based on a given natural language instruction.
First, we extract all noun-phrases (NPs) from instructions in the dataset and compute their~\textsc{Clip} language embedding.
Given a visual observation, we compute~\textsc{Clip} visual embedding for each image patch, and align it to the NP with the highest cosine similarity between~\textsc{Clip} embeddings.
We then combine aligned NPs with agent orientation to formulate an egocentric language description of the scene.
This is described in more detail in Appendix~\ref{app:touchdown_verbalization}.

\subsection{Learning a feedback model}
\label{sec:learning_feedback_model}

\paragraph{Naively learning from LLM feedback.}
Given a verbalization procedure $\verbalize$, an instruction $x$, an LLM, and a policy $\policy_\param$, we now describe a procedure to use the LLM's knowledge to improve $\policy_\param$.
First, we prompt the LLM to provide feedback on whether a particular action taken by the policy $\policy_\param(a|x, \verbalized)$ is productive in achieving the tasks outlined in the instruction $x$.
We then improve the policy $\policy_\theta$ by updating its parameters to imitate desirable behaviour determined by the LLM.
Let $:$ denote ``such that''.
Let $\llmmethod(x, \verbalized, a)$ return $\true$ if and only if the LLM feedback indicates that action $\action$ taken in verbalized state $\verbalized$ and instruction $\instruction$ is productive.
Given a set of instructions $X = \{x_i\}_1^N$, the optimization procedure is then
$
    \label{eq:naive_feedback}
    \argmin_\theta \expect_{
    \verbalized, \action^\prime, \instruction: \llmmethod(\instruction, \verbalized, \action^\prime) = \true
    } \loss \left( \policy_\theta \left( \action|\instruction, \verbalized \right), \action^\prime \right)
$.
Here, instruction $\instruction$ is sampled from $X$.
Observations $\verbalized$ and actions $\action^\prime$ are sampled from rollouts of the policy $\policy_\theta$.

\paragraph{Efficiently learning a language feedback model.}
While the previously described naive learning is a reasonable procedure for using LLM feedback to improve the policy, it requires calling LLMs at each step during policy improvement.
This is prohibitively expensive both in terms of query cost, because LLMs capable of giving desirable feedback are expensive to run, and training time, because generating feedback using large LLMs is slow.
Instead of using the LLM at each step, we make a modification to collect LLM feedback over long horizons in batch~\cite{pmlr-v232-colas23a} in order to train a small and cost-effective language feedback model.

First, for instructions $\{\instruction^{(1)}, \instruction^{(2)},\ldots\}$ we roll out the base policy $\policy_\theta$ to collect a set of trajectories $\{\traj_1, \traj_2, \ldots\}$ consisting of verbalized observations and actions taken: $\traj_i = \{ \verbalized^{(i)}_1 \policy (\instruction^{(i)}, \verbalized^{(i)}_1), \verbalized^{(i)}_2 \policy (\instruction^{(i)}, \verbalized^{(i)}_2), \ldots \}$.
For each $\traj_i$, we prompt the LLM for feedback on which steps were productive in achieving the instruction $\instruction^{(i)}$.
Table~\ref{tab:prompts}'s~\methodOurs~row shows an example of requesting feedback from GPT-4 on a rollout in ALFWorld, which is an instruction following benchmark in verbalized 3D kitchens.
This LLM feedback is then parsed to identify the precise steps in which the base policy $\policy_\theta$ took a productive action towards achieving the goals outlined in the instruction.
The set of desirable behaviour is compiled into a dataset $\feedbackdataset$.
Let $\feedbacklabel = \llmmethod(\instruction, \verbalized, \action)$ denote the feedback given by the LLM for the instructions $x$, observations $\verbalized$, and action $\action$.
We use the dataset $\feedbackdataset = \{\instruction^{(i)}, \verbalized, \action, \feedbacklabel \forall \verbalized, \action \in \traj_i \forall \instruction^{(i)}, \traj_i \}$ to train a small~\methodname~$\feedbackmodel$:
\begin{equation}
    \argmin_\theta \sum_{(\instruction, \verbalized, \action, \feedbacklabel) \in \feedbackdataset}
    \loss \left(
        \feedbackmodel_\theta \left(\feedback \mid \instruction, \verbalized, \action \right)
        ,
        \feedbacklabel
    \right).
\end{equation}
Here, $\loss$ is the cross-entropy between the feedback model output $\feedbackmodel_\theta$ and gold label $\feedbacklabel$ from the LLM.

\paragraph{Learning from language feedback.}
The naive learning procedure in Eq~\eqref{eq:naive_feedback} updates the policy after each step using slow and expensive LLM feedback.
Here, we instead update the policy in rounds using fast and cost-effective~\methodnameshort~feedback.
In round $k$, we rollout the base policy $\policy^{(k)}$ and use the feedback model $\feedbackmodel$ to collect a dataset $\dataset_k$ of desirable behaviour.
Let $a^{(k)}_t$ denote the action chosen by policy $\policy^{(k)}(a \mid \instruction, \verbalized_t)$.
Let $\desirable(\instruction, \verbalized, \action) = \feedbackmodel \left( \feedback = \true \mid \instruction, \verbalized, \action \right) > \feedbackmodel \left( \feedback = \false \mid \instruction, \verbalized, \action \right)$ return whether the feedback model predicts that action $\action$ is desirable.
In the $k$th round, we collect the dataset of desirable behaviour
%
$
\dataset_k = \left\{
        \left( \instruction, \verbalized_t, \action^{(k)}_t \right)
        \forall t
        : \desirable(\instruction, \verbalized_t, \action^{(k)}_t)
\right\}
$,
%
which we combine with previously collected behaviour to update the policy via imitation learning:
\begin{eqnarray}
    \label{eq:training}
    \theta^* = \argmin_\theta \sum_{v_t, a_t \in \cup_{i=1}^k \dataset_i} \loss \left(
    \policy^{(k)}(a \mid \instruction, \verbalized_t)
    , a_t \right).
\end{eqnarray}
In the next round, we set the base policy $\policy^{(k+1)}$ parameters to $\theta^*$.
Should demonstrations be available, we initialize the base policy at $k=1$ to the BC policy, and train on both demonstrations and identified desirable behaviour during subsequent rounds (i.e.~$\cup_{i=0}^k \dataset_i$ where $\dataset_0$ are demos used to train BC).

\section{Related Work}

\paragraph{Instruction following in grounded environments.}
Instruction following in grounded environments has been explored in settings such as navigation~\citep{chenLearningInterpretNatural2011,friedUnifiedPragmaticModels2018,chenTouchdownNaturalLanguage2019}, game-playing~\citep{andreasAlignmentbasedCompositionalSemantics2015,zhongRTFMGeneralisingNovel2020}, and robotics~\citep{blukisLearningMapNatural2019,shridharCLIPortWhatWhere2021,brohanRT2VisionLanguageActionModels2023}.
However, most prior work model environment observations separately from language instructions by using specialized encoders (e.g.~\textsc{ResNet}~\citep{heDeepResidualLearning2015},~\textsc{Bert}~\citep{devlinBERTPretrainingDeep2019}, ~\textsc{Clip}~\citep{radfordLearningTransferableVisual2021}), then learn from data how to associate raw observations with language instructions.
Instead of solely using raw observations, more recent work verbalize raw observations to describe environments in language~\citep{ALFWorld20,zhongSILGMultienvironmentSymbolic2021,schumannVELMAVerbalizationEmbodiment2024}.
In doing so, observations and instructions can be directly jointly reasoned over using language models to achieve more efficient and generalizable learning through large-scale pretraining.
We build on this last direction by verbalizing raw observations into language descriptions to train language policies.
However, unlike prior work that train language models to predict next actions, we develop language feedback models that critique verbalized observations and behaviour.

\paragraph{LLM agents in language settings.} 
LLMs exhibit reasoning abilities after pretraining on vast quantities of text~\citep{brownLanguageModelsAre2020,weiChainofThoughtPromptingElicits2022}.
A number of recent work on LLMs language agents exploit this reasoning ability.
\citet{nakanoWebGPTBrowserassistedQuestionanswering2022},~\citet{yaoReActSynergizingReasoning2023}~\citet{dengMind2WebGeneralistAgent2023}~train instruction following language agents to interact with web browsers to answer questions or interact with web pages.
\citet{ahnCanNotSay2022}~show that a language agent can be connected with verbalized robots via API interfaces for robotic control.
\citet{Xie2024OsworldBenchmarkingMultimodal}~use large visual language models (VLMs) for instruction following in virtual machines, but show that LLMs with verbalization outperform VLMs.
While powerful, these prior work are limited in that they require querying an expensive LLM on-line.
In contrast, our work examines settings where an LLM is not available on-line.
Specially, we use LLMs to collect a small set of off-line data for training~\methodnameshort s.
The small and cost-effective~\methodnameshort s are then used to identified desirable behaviour for on-line policy improvement without additional interactions with the LLM.

\begin{table*}[!t]
    \small
    \centering
    \caption{Examples of verbalization. We abbreviate long verbalized observations using ``...''.}
    \label{tab:verbalized_environments}
    \begin{tabularx}{\textwidth}{l X l}
        \toprule
        Benchmark & Context & Action \\
        \midrule
        ALFWorld & 
        \makecell[{{p{9cm}}}]{
        Task: heat some egg and put it in diningtable.\\
        Observation: You arrive at loc 12. On the sinkbasin 1, you see...\\
        T-1 Observation: You are in the middle... Action: go to sinkbasin 1\\
        T-2 Observation: ...
        }
        & \makecell{go to\\microwave 1}\\
        \midrule
        ScienceWorld &
        \makecell[{{p{9.5cm}}}]{
        Task: Your task is to find a(n) living thing. First, focus on the thing. Then, move it to the purple box in the bathroom.\\
        Observation: You move to the kitchen. This room is called the kitchen. In it, you see:
        | the agent | a substance called air | a chair. On the chair is...\\
        In your inventory, you see: | an orange...\\
        T-1 Observation: The door is now open. Action: go to kitchen\\
        T-2 Observation... Action: open door to kitchen
        }
        & \makecell{open door\\to outside}\\
        \midrule
        Touchdown &
        \makecell[{{p{9.5cm}}}]{
        Task: Follow the flow of traffic, with the row of flowers on your left and make a left at
        the intersection. There will be a white billboard...\\
        Observation: behind you, you see: the right lane intersection, a large...\\
        T-1 Observation: behind you, slightly... Action: slightly to your left
        ...
        } & \makecell{straight\\ahead} \\
        \bottomrule
    \end{tabularx}
    \vspace{-0.2in}
\end{table*}

\paragraph{Learning from feedback.}
Recent work enhance language agents by augmenting them with feedback.
\citet{zieglerFineTuningLanguageModels2020},~\citet{stiennonLearningSummarizeHuman2020}, and~\citet{baiConstitutionalAIHarmlessness2022} learn reward models from human preference to improve policies via reinforcement learning (RL).
Instead of using human feedback,~\citet{baiConstitutionalAIHarmlessness2022}~and~\citet{leeRLAIFScalingReinforcement2023}~use LLM feedback to train a separate reward model for RL for textual alignment.
\citet{huangInnerMonologueEmbodied2022},~\citet{yaoReActSynergizingReasoning2023}, and~\citet{Shinn2023-reflexionlearning}~use LLMs to reason about potential resolutions to failed actions.
\citet{yuanSelfRewardingLanguageModels2024}~use LLMs to generate new prompts and corresponding responses, then use an LLM reward model to identify good prompt-response pairs for self-improvement in text generation alignment.
Unlike these approaches, we do not use LLMs during on-line policy improvement.
We train an initial small language feedback model from offline LLM data, then use this small feedback model for policy improvement.
Additionally, we focus on-line improvement via language feedback for long-horizon, sparse reward, grounded environments instead of text generation alignment. 
Our procedure for batched, on-line imitation learning is similar to~\textsc{Dagger}~\citep{rossReductionImitationLearning2011}, which we compare to in Appendix~\ref{app:dagger}.
However, we collect batched expert feedback to identify desirable behaviour instead of corrective actions.
\citet{klissarovMotifIntrinsicMotivation2023}~and \citet{duGuidingPretrainingReinforcement2023}~are recent works that describe learning from feedback approaches complementary to ours.
The former learns preference models based on pairwise observations while the latter uses LLMs to suggest exploratory goals during training.
Unlike these works, which assume that the underlying goal is the same between training and inference, we consider settings where training and evaluation goals are different.
That said, one can expand these approaches to generalize to unseen environments by adapting a preference model during inference~\citep{klissarovMotifIntrinsicMotivation2023} and by goal-conditioned subgoal generation during inference~\citep{duGuidingPretrainingReinforcement2023}.
However, unlike~\methodnameshort, these modifications would then rely on calling LLMs during inference.

\section{Experiments and Analysis}

We evaluate using~\methodnameshort~s for policy improvement on three distinct language grounding benchmarks.
Formally, the~\textbf{environment}s from a~\textbf{benchmark} are distinct partially-observed Markov Decision Processes that share some (or all) of the environment dynamics but have different instructions, observations, and/or action space. 

\subsection{Evaluation benchmarks}
Table~\ref{tab:verbalized_environments}~shows examples of verbalized environments and tasks from each benchmark.
Each benchmark provides distinct training and test environments to test generalization.
In each environment, the agent takes actions to perform tasks outlined in a language instruction.
The task is considered completed if and only if the agent solves the tasks within the preallocated number of steps.
We evaluate using task-completion rate over test environments.
The statistics from each benchmark is shown in Appendix~\ref{app:touchdown_verbalization} Table~\ref{tab:benchmark_stats}.
These three benchmarks share challenges in sparse, delayed reward, partial observability, and compositional generalization to unseen tasks and environments.

\textbf{ALFWorld} is a verbalization of ALFRED~\cite{ALFRED20}, a natural language instruction following benchmark set in a 3D simulated kitchen.
Here, the agent interacts with objects in kitchens to achieve compositional goals such as cleaning then microwaving potatoes.
In ALFWorld~\cite{ALFWorld20}, raw state information from ALFRED are used to populate language templates that describe observations in language. 

\textbf{ScienceWorld} is a textual simulation benchmark for basic science experiments~\cite{Wang2022ScienceWorld}.
The agent interacts with objects to conduct experiments specified in natural language, such as determining the boiling temperature of a material.
ScienceWorld is uniquely challenging to due the large amount of variations in task types (30), and parametric variations (10-1400) such as the specific substance to be melted.
Furthermore, ScienceWorld has a substantially larger action space and longer horizon tasks.

\textbf{Touchdown} is a navigation benchmark where the agent navigates Google Street View images to follow long, compositional instructions~\cite{chenTouchdownNaturalLanguage2019}.
Touchdown requires jointly reasoning over natural images from Google Streetview with occlusion and multi-sentence natural language instructions that describe long-horizon goals.
We introduce a new verbalization procedure for Touchdown based on matching noun-phrases and image patches with~\textsc{Clip} embeddings to populate egocentric language templates.
Behaviour cloning using our verbalization is detailed in Appendix~\ref{app:touchdown_verbalization}.
Touchdown considers multiple subtasks, in this work we only test the agent's ability to arrive at the correct location according to the instruction.

\subsection{Methods}

\begin{table*}[!t]
    \small
    \centering
    \caption{LLM prompts used to collect desirable behaviour.~\methodActionPrediction~uses LLMs to directly generate actions for each step, whereas~\methodOurs~uses LLMs to generate batch feedback that identify which taken actions were productive. For brevity, we abbreviate long verbalized observations using ``...''. ``Before'' contains the observation before the first step in the batch.}
    \label{tab:prompts}
    \begin{tabularx}{\textwidth}{l X l}
        \toprule
        \multicolumn{2}{c}{\textbf{\methodActionPrediction}}\\
        \midrule
        \rowcolor{lightgray}
        \textbf{Prompt} & \makecell{Your task is: look at alarmclock under the desklamp.\\
        You see: you are in the middle of a room. looking quickly around you, you see a bed 1...\\
        what do you decide to do? available actions: examine shelf 1, examine shelf 2, go to bed...\\
        You decide to: go to desk 1.\\
        You see: you arrive at desk 1.
        what do you decide to do? available actions: examine desk...\\
        You decide to:}\\
        \rowcolor{lightergreen}
        \textbf{LLM Output} & examine desk 1\\
        \midrule
        \multicolumn{2}{c}{\textbf{\methodOurs}}\\
        \midrule
        \rowcolor{lightgray}
         \textbf{Prompt} & \makecell{You will be shown a playthrough for solving a task.\\
        Task: put two candle in drawer.\\
        Before: You open the drawer 6. The drawer 6 is open. In it, you see nothing.\\
        Step 27. Your action: close drawer 6. Result: You close the drawer 6...\\
        Step 28. Your action: put candle 3 in/on drawer 1. Result: You put the candle 3 in...\\
        Is the player on the right track to solve the task?\\
        Answer yes or no. If yes, list the helpful steps by the step number in bullet form.}\\
        \rowcolor{lightergreen}
        \textbf{LLM Output} & 
        \makecell{Yes\\- Step 28\\- Step 29...}\\
        \bottomrule
    \end{tabularx}
    \vspace{-0.2in}
\end{table*}

We train BC baseline policies using existing demonstrations.
We examine three different techniques for improving the BC policy.
Table~\ref{tab:prompts}~shows examples of LLM prompts used for each technique.

\paragraph{\methodActionPrediction: imitation learning from LLM experts.}
We compare to directly using LLMs as experts to predict actions for imitation learning.
First, we execute $k$ steps of the base policy, then query the LLM for the next action $\action$ given the instruction $\instruction$ and the verbalized observations $\verbalized$.
We repeatedly collect examples $(\instruction, \verbalized, \action)$, then train the policy using this collected data and BC demonstrations.

\paragraph{\methodOurs: imitation learning using feedback models.}
We learn a small and cost-effective feedback model described in Section~\ref{sec:learning_feedback_model} to identify desirable behaviour for imitation learning.
First, we learn a feedback model on the training environments.
Second, we use the feedback model to identify desirable behaviour in the training environments for policy improvement via imitation learning.
To collect LLM feedback for training~\methodnameshort s, we collect one rollout for each environment in a benchmark and sample 10k 20-step windows from the rollouts.
Crucially, we limit the amount of feedback data collected from the LLM such that the number of output tokens produced by the LLM is identical to~\methodActionPrediction~(we use 100k GPT-2 tokens for all benchmarks).
For~\methodnameshort, we collect feedback for as many windows as possible until we exceed 100k output tokens, then use this feedback to train the LFM.
For~\methodActionPrediction, we label actions until we exceed 100k output tokens, then combine this labeled set with demonstrations to train the policy. 
This limitation on LLM interactions answers whether the feedback model is more cost-effective than direct action prediction for imitation learning.

\paragraph{\methodOursAdapt: adaptation using feedback models.}
\methodOurs~only imitates desirable behaviour in training environments.
In contrast,~\methodOursAdapt~adapts the policy to test environments.
Given new test environments, we identify desirable behaviour using feedback models trained on the training environments, then perform one round of imitation learning to adapt to new test environments.
This experiment tests whether language feedback models generalize to new environments, and whether we can use their feedback to adapt policies to new environments without using LLMs nor additional demonstrations.

\begin{wraptable}{r}{0.55\linewidth}
    \vspace{-0.2in}
    \centering
    \small
    \caption{
    Task completion rates of behaviour cloning~\methodBehaviourCloning, imitation learning (IL) using LLM expert~\methodActionPrediction, and IL using~\methodOurs.
    On held-out test environments, \methodOurs~outperforms other methods on all benchmarks.
    \methodActionPrediction~and~\methodOurs~are limited to 100k output tokens of GPT-4 interactions.
    Further adaptation to the new environments using~\methodnameshort~results in significant additional gains (\methodOursAdapt).
    Errors are standard deviations across 3 seeds.
    Previous SOTA are~\citet{micheliLanguageModelsAre2021} for ALFWorld,~\citet{linSwiftSageGenerativeAgent2023} for ScienceWorld, and~\citet{schumannAnalyzingGeneralizationVision2022} for Touchdown.
    Unlike~\citet{linSwiftSageGenerativeAgent2023}, our methods do not use ScienceWorld-specific custom room tracking nor action reranking.
    }
    \label{tab:main_results}
    \begin{tabularx}{\linewidth}{cXXX}
    \toprule
         & ALF & SciWorld & TD \\
    \midrule
    Prev SOTA & 57.6 & 45.8 & 29.3 \\
    \midrule
    GPT-4 zeroshot & 3.0 & 1.3 & 3.2 \\
    \midrule
    \methodBehaviourCloning    &  $62.6 \pm 0.4$&  $45.8 \pm 0.6$& $57.5 \pm 0.3$\\
    \methodActionPrediction    &  $56.0 \pm 0.7$& $39.0 \pm 0.7$& $58.0 \pm 0.4$\\
    \methodOurs    &  $\textbf{64.1} \pm 0.3$& $\textbf{47.1} \pm 0.5$& $\textbf{59.7} \pm 0.4$ \\
    \midrule
    \methodOursAdapt~1 rnd &  $74.6 \pm 1.1$& $49.3 \pm 0.9$& $62.8 \pm 1.1$\\
    \methodOursAdapt~2 rnds  &  $76.5 \pm 1.3$& $50.4 \pm 1.0$& $63.5 \pm 1.2$\\
    \bottomrule
    \end{tabularx}
    \vspace{-0.40in}
\end{wraptable}

\subsection{Experiment details}
We use the GPT-4 (\texttt{2023-03-15}) for action prediction and feedback, and finetune 770M~\textsc{Flan-T5}~\citep{chungScalingInstructionFinetunedLanguage2022} for policy and feedback models.
Verbalized observations $\verbalized$ contain the most recent 20 steps.
We train models for 10k steps with batch 20, learning rate 5e-5, and early stopping over validation demos.
For~\methodActionPrediction~and~\methodnameshort, we limit the amount of LLM usage to 100k GPT-2 tokens.
Touchdown verbalization uses~\texttt{vit-large-patch14}.
Appendix~\ref{app:devices} details GPU usage.

\paragraph{Feedback model training and inference.}
We train~\methodnameshort s using LLM feedback over 20-step windows.
We then parse feedback to identify whether the action taken in each step was productive to solving the tasks outlined in the instructions.
We subsample the feedback data to obtain an even split of productive and not-productive actions.
This data is split into a 80\% train/20\% validation dataset to train the~\methodnameshort.

\paragraph{Policy training and inference.}
To train policies, we fine-tune language models to minimize token-wise cross-entropy of the ground-truth verbalized action.
During inference, we consider a (potentially very large) set of plausible actions given by the environment.
For each action, we evaluate the policy's language model perplexity, and choose the action with the minimum perplexity averaged over tokens.

\subsection{Results and discussion}
\label{sec:results}

Table~\ref{tab:main_results} shows the performance of the policy behaviour cloned from demonstrations~\methodBehaviourCloning, imitation learned from LLMs using action prediction~\methodActionPrediction, and imitation learned from~\methodOurs.
For~\methodnameshort s, we show zero-shot (\methodOurs) as well as adaptation (\methodOursAdapt) results.

\paragraph{\methodnameshort s~improves policy performance across all benchmarks.}
Table~\ref{tab:main_results} shows that~\methodOurs~improves upon the strong behaviour cloning baseline policy~\methodBehaviourCloning~in all benchmarks.
Table~\ref{tab:detailed_feedback} shows examples of~\methodnameshort-identified desirable behaviour.
This shows that~\methodnameshort s are an effective means to leverage the knowledge in pretrained LLMs for policy improvement in language-grounded environments, which agree with human-identified desirable behaviour.
Appendix~\ref{app:llama} also compares GPT-4 to the open-source~\textsc{Llama 2} 70B for training feedback models using human evaluation.
We find that GPT-4 consistently outperforms~\textsc{Llama 2}, which tends to identify spurious desirable behaviour.

\paragraph{Learning~\methodnameshort s is more cost-effective than using LLMs for action prediction.}
Assuming the same LLM output-token quota, Table~\ref{tab:main_results} compares using LLMs to train feedback models (\methodOurs) to using LLMs to predict actions (\methodActionPrediction) for policy improvement.
Specifically,~\methodActionPrediction~tends to predict spurious actions, especially for complex environments with large actions spaces such as ScienceWorld.
In contrast, the difficulty in identifying productive actions is independent of the action space, and~\methodOurs~consistently improves policy even with large action spaces.
This shows that~\methodnameshort s is a more cost-effective means use LLMs for policy improvement compared to using LLMs as expert policies for imitation learning.

\paragraph{\methodnameshort s generalize to new environments, allowing for policy adaptation without additional LLM usage nor demonstrations.}
Table~\ref{tab:feedback_results}~shows that~\methodnameshort s trained during language feedback learning can accurately recognize desirable behaviour in new environments.
Table~\ref{tab:main_results}~shows that imitating this behaviour obtains significant policy improvement across all benchmarks.
This shows that~\methodnameshort s generalize to new environments, which allows for policy adaptation to new environments despite not having demonstrations nor LLM access.
Appendix~\ref{sec:correlation}~shows additional experiments that demonstrate correlation between LLM/\methodnameshort~feedback and true state values.

\begin{wraptable}{r}{0.4\linewidth}
    \vspace{-0.2in}
    \centering
    \caption{
    Feedback performance of~\methodnameshort.
    We measure F1 score of the productive/not-productive predictions made by the learned~\methodnameshort~using the LLM predictions as ground truth.
    We observe no significant performance degradation when using a much more detailed feedback model (\methodOursDetailed) that also provides explanations behind the feedback, summaries of agent behaviour, and strategy suggestions.
    }
    \label{tab:feedback_results}
    \begin{tabularx}{\linewidth}{cXXX}
    \toprule
         & ALF & SciWorld & TD \\
    \midrule
    \methodOurs & 93.2  & 83.7 & 43.9 \\
    \methodOursDetailed & 92.0 & 82.5 & 42.5 \\
    \bottomrule
    \end{tabularx}
    \vspace{-0.2in}
\end{wraptable} 

\begin{table}[!t]
    \small
    \centering
    \caption{Example of detailed human-interpretable feedback.
    The prompt and output differences between succinct feedback (Table~\ref{tab:prompts}) and detailed feedback are highlighted in~\textcolor{darkred}{red}.
    Note that the second row shows shows batched LLM inference, which is used to train a step-wise feedback model.
    The step-wise inference of the feedback model is shown in the last row.
    We leave providing summaries and suggestions with the learned feedback model to future work.
    }
    \label{tab:detailed_feedback}
    \begin{tabularx}{\linewidth}{X}
    \toprule
    \rowcolor{lightgray}
    \makecell{
    \textbf{Prompt} \\
    You will be shown a playthrough for solving a task.
    Task: put two candle in drawer.
    Before: You open the\\ drawer 6. The drawer 6 is...
    Step 21. Your action: close drawer 6. Result: You ...
    Is the player on the right\\track to solve the task?\\
    \textcolor{darkred}{Next under heading \#Summary, summarize in one sentence what the player doing at a high level. Finally}\\
    \textcolor{darkred}{under heading \#Improvement, describe how the player can improve their strategy to solve the task.}
    }
    \\
    \midrule
    \rowcolor{lightergreen}
    \makecell{
    \textbf{LLM output} \\
    Yes\\
    - Step 28: \textcolor{darkred}{The player locates the candles by arriving at countertop 1, where a candle is available.}\\
    - Step 29: \textcolor{darkred}{The player successfully takes candle 3 from countertop 1, demonstrating the ability to collect...}\\
    \textcolor{darkred}{\#Summary}\\
    \textcolor{darkred}{The player is gathering candles and placing them in a drawer, working towards completing the task of...}\\
    \textcolor{darkred}{\#Improvement}\\
    \textcolor{darkred}{The player can improve their strategy by: not closing drawers unnecessarily...}\\
    } \\
    \rowcolor{lightergreen}
    \makecell{
    \textbf{Learned feedback model output}\\
    Yes. \textcolor{darkred}{The player successfully locates the candle...}\\
    Yes. \textcolor{darkred}{The player correctly takes the candle from...}\\
    } \\
    \bottomrule
    \end{tabularx}
    \vspace{-0.25in}
\end{table}

\paragraph{\methodnameshort s can provide human-interpretable feedback, allowing human-in-the-loop verification during policy improvement.}
\methodnameshort s~improve policy performance with succinct feedback.
Here, we extend them to additionally provide detailed explanations.
Consider an instruction ``turn left when you see the stop sign then go to the second building on the right''.
Suppose that in the current step the agent proceeds straight, arriving at the stop sign.
Instead of a feedback saying ``yes'' (i.e.~the action was productive), the~\methodnameshort~can provide a human-interpretable explanation for why this action was productive (i.e. ``yes because you found the stop sign where you are supposed to turn'').
Table~\ref{tab:detailed_feedback}~shows that we can enhance~\methodnameshort~to produce detailed feedback by training detailed feedback prompted from LLMs.
Specifically, we train a detailed~\methodOursDetailed~to simultaneously identify productive actions, summarize agent intent, and suggest potential high level recovery strategies.
Table~\ref{tab:feedback_results}~shows that surprisingly,~\methodOursDetailed~that produce detailed feedback perform similarly to those that provide succinct feedback.  
This shows that~\methodname s can be used to provide accurate feedback interpretable to humans.
While interpretable feedback requires more costly LLM usage, it allow for human-in-the loop verification of desirable behaviour identified by the~\methodnameshort.
Consequently, interpretable~\methodnameshort s promotes user trust in the quality of the imitation learning data and subsequent policy behaviour.

\section{Conclusion}
We introduced~\methodname s that identify desirable behaviour for imitation learning.
On three instruction following benchmarks, small and cost-effective~\methodnameshort s consistently outperform BC baselines and using LLMs as experts for imitation learning, without using LLMs during policy improvement.
In addition,~\methodnameshort s generalize and provide significant policy adaptation gains on new environments, without using LLMs nor new demonstrations.
Finally,~\methodnameshort s, can provide detailed human-interpretable feedback that human verification of imitation data.
We advocate for future exploration of how to exploit detailed~\methodnameshort s, such as learning dense, subgoal-aware reward models for RL, and trustworthy policies with human verification.

\bibliography{neurips_2024}
\bibliographystyle{abbrvnat}


\appendix
\newpage
\appendix
\onecolumn

\section{Limitations}
\label{app:limitations}

This work proposes using LLM feedback to improve policies for long-horizon planning in grounded environments.
It assumes access to a verbalization module that faithfully describes observations in language.
For some practical problems (e.g.~rich scenes in operating systems~\citet{xieText2RewardAutomatedDense2024}), verbalization has been shown to be a difficult problem, which limits the capability of language feedback models.
Moveover, LLMs have been shown to hallucinate, especially in grounded settings that are uncharacteristic in internet pretraining data~\citep{Wang2022ScienceWorld}.
Consequently, LLMs may provide inaccurate feedback, which limits policy improvement gains.

Our empirical results demonstrate consistent policy improvement using~\methodnameshort s on three distinct language grounding benchmarks.
These benchmarks cover kitchen interactions (AFLWorld), scientific experiments (ScienceWorld), and real-scene navigation (Touchdown).
Future work should investigate the application of LLM feedback learning to real-world robotics, as an important test-bed for LLM ability to provide precise feedback on real-world observations.

While this work significantly reduces the computational burden of using language models as policies in grounded environments by extracting world knowledge from LLMs into small LMs using language feedback, the LM policies we use still require large GPUs to train.
Specifically, we note significant performance degradation (e.g.~10\% task completion) when changing from 770M Flan-T5 to the smaller 250M variant.
Future work should investigate techniques to further reduce model size so that LMs can be tractably used as policies on small devices such as phones.

\section{Broader Impacts}
\label{app:impact}

Although this work develops LM as policies in grounded environments, the outputs these policies generate are specific (i.e.~plans) to the benchmark environments used.
Consequently, the generated outputs are not suitable for malicious use (e.g.~disinformation, fake profiles).
One potential misuse of our proposed method lies in a central party providing a malicious feedback model, which provides negative reinforcement to good behaviour and positive reinforcement to bad behaviour.
Learning from such a feedback model can potentially result in malicious downstream policies.

\section{Licenses for existing assets}
\label{app:licenses}

We will release our code and resources under the MIT license.
This work uses assets from the Flan-T5 series of models, OpenAI GPT-4, Llama-2, and the three benchmarks.
The licenses are as follows:

\begin{itemize}
    \item Flan-T5: \href{https://github.com/google-research/FLAN/blob/main/LICENSE}{Apache 2}
    \item OpenAI GPT-4: \href{https://openai.com/policies/usage-policies/}{fair use policy}
    \item Llama-2: \href{https://ai.meta.com/llama/license/}{community license}
    \item ALFWorld: \href{https://github.com/alfworld/alfworld/blob/master/LICENSE}{MIT}
    \item ScienceWorld: \href{https://github.com/allenai/ScienceWorld/blob/main/LICENSE}{Apache 2}
    \item Touchdown: \href{https://github.com/lil-lab/touchdown/blob/master/LICENSE.txt}{Attribution 4 International}
\end{itemize}

\newpage
\section{Verbalization of visual environments}
\label{app:touchdown_verbalization}

\begin{table}[!t]
    \centering
    \caption{
    Statistics from benchmarks as measured by training demonstrations.
    The are the average number of GPT-2 tokens in the instruction, verbalized observation, and action; the average demonstration steps; the average number of plausible actions in a state; the number of unique actions, instructions, and observations; and finally the number of training demonstrations.
    }
    \label{tab:benchmark_stats}
    \begin{tabularx}{\linewidth}{XXXXX}
    \toprule
    & ALFWorld & SciWorld & Touchdown\\
    \midrule
    Ins len $|\instruction|$    & 8.8   & 64.7  & 93.4\\
    Obs len $|\verbalized|$     & 23.9  & 239.4 & 284.9\\
    Act len $|\action|$         & 4.5   & 6.0   & 2.4\\
    \midrule
    Traj len $|\traj|$          & 19.7  & 55.1  & 34.2\\
    \midrule
    $|$Act space$|$             & 29.9  & 1.9k  & 2.1\\
    \midrule
    \# act $|\{\action\}|$      & 2.6k  & 2.4k  & 8\\
    \# ins $|\{\traj\}|$        & 1.0k  & 1.2k  & 6.5k\\
    \# obs $|\{\verbalized\}|$        & 18.2k & 157k  & 34.3k\\
    \# demos                    & 3.5k  & 3.6k  & 6.5k\\
    \bottomrule
    \end{tabularx}
    \vspace{-0.1in}
\end{table}

How can we leverage the world knowledge learned by LLMs from pretraining on vast quantities of text?
In many instruction following problems, environment observations are inherently visual.
In this section, we describe a verbalization procedure that converts visual observations to language descriptions, so that LLMs can make inferences by jointly referring to the instruction and environment observations.
Specifically, we use Touchdown as an example.

As shown in Figure~\ref{fig:touchdown_verbalization}, Touchdown~\citep{chenLearningInterpretNatural2011}~is primarily a visual navigation problem.
Given a set of connected Google Streetview panoramas that represent neighbourhoods in Manhattan, an agent must follow long, complex natural language instructions to navigate to the correct location.
Crucial to this task of navigation are~\textbf{landmarks} referred to by the instructions.
For instance, the instruction ``turn so the~\textbf{scaffolding} is on your left and... to the~\textbf{next corner} and turn right...'' refers to the landmarks~\textbf{scaffolding} and~\textbf{next corner}.
Prior work in verbalization use LLMs to identify landmarks~\citep{schumannVELMAVerbalizationEmbodiment2024}.
In this work, we take the set of common noun-phrases in the corpus of instructions to be landmarks.

\paragraph{Extracting aligned noun-phrase annotations for visual patches}
First, we identify all noun-phrases using \textsc{SpaCy}~\citep{spacy2}.
Given a visual scene, we divide the scene into 300x300 pixel non-overlapping patches.
For each patch, we identify the noun-phrase with the highest cosine similarity between the noun-phrase text embedding and the image patch embedding.
We use text and visual encoders from \textsc{Clip}~\citep{radfordLearningTransferableVisual2021} to extract embeddings for each modality.
For patches with no aligned noun-phrase with cosine similarity greater than 0.2, we do not provide annotated a noun-phrase.

\paragraph{Converting to verbalized observations}
To obtain verbalized observations in the form of an egocentric scene description, we consider the direction the agent is facing (shown in~\textcolor{darkblue}{blue}) as well the directions of possible next steps (shown in~\textcolor{darkgreen}{green}).
The noun-phrases identified in the scene are then categorized into 8 directions in 45-degree increments, relative to the agent's current orientation: straight ahead (337.5 to 22.5), slightly to your right (22.5 to 67.5), to your left (67.5 to 112.5), behind you, slightly to your right (112.5 to 157.5), behind you (157.5 to 202.5), behind you, slightly to your left (202.5 to 247.5), to your left (247.5 to 292.5), and slightly to your left (292.5 to 337.5).
A scene is then rendered as follows:
{\small
\begin{verbatim}
Straight ahead, you see a white van
Slightly to your right, you see a red brick building, a scaffold...
\end{verbatim}
}

It is important to note that Touchdown consists of multiple subtasks, such as finding the correct scene, stopping at the correct scene, and then finally orienting to face a hidden object in the scene.
In this work, similar to~\citet{zhongSILGMultienvironmentSymbolic2021}, we only consider the first task of finding the correct scene according to the instruction.
To compare our verbalization with prior work, we also evaluate a separate setting (e.g.~used in~\citet{chenTouchdownNaturalLanguage2019,schumannVELMAVerbalizationEmbodiment2024}) where the agent must identify when to stop and is credited so long as it stops within one panorama of the target scene.
Behaviour cloning using our verbalization technique results in 63.0\% task-completion rate.

\paragraph{Statistics of verbalized environments}
In Appendix~\ref{app:touchdown_verbalization}~Table~\ref{tab:benchmark_stats}, we show statistics of verbalized environments as quantitative evidence of their challenges.

\newpage
\section{Pseudocode for Policy Improvement using~\methodname s}
\label{app:pseudocode}

In this section we detail, using pseudocode, the procedure for policy improvement using~\methodname s.
Algorithm~\ref{alg:train_feedback_model}~describes learning a model from LLMs.
Algorithm~\ref{alg:imitating_desirable_behaviour}~describes identifying desirable behaviour that are productive for solving tasks specified in the instruction, and then using this behaviour for imitation learning.
Algorithm~\ref{alg:improve_policy} describes the iterative policy improvement procedure using these two algorithms.

\begin{algorithm}
\caption{\textsc{TrainFeedbackModel}: Training a~\methodname~using LLM feedback.}
\label{alg:train_feedback_model}
\begin{algorithmic}[1]
\STATE Inputs: initial policy $\policy$, LLM $\llmmethod$, environment $\env$
\STATE Feedback dataset $\feedbackdataset \gets \{\}$
\FOR{$i = 1 \ldots N$}
    \STATE $\instruction \gets \mathsc{SampleInstruction}$
    \STATE $\traj_i \gets \rollout(\policy, \env, \instruction)$
    \FOR{window $w_j$ in $\traj_i$}
        \STATE $y \gets \mathsc{QueryLLMForFeedback}(\llmmethod, w_j, \instruction)$
        \FOR{verbalized observation $\verbalized_k$, LLM feedback $y_k$ in each step of $y$}
            \STATE $\feedbackdataset \gets \feedbackdataset \bigcup (\verbalized_k, y_k)$
        \ENDFOR
    \ENDFOR
\ENDFOR

\STATE Feedback model $\feedbackmodel \gets \mathsc{TrainLM}(\feedbackdataset)$
\end{algorithmic}
\end{algorithm}

\begin{algorithm}
\caption{\textsc{ImitateUsingFeedback}: Imitation learning using desirable behaviour identified by a feedback model.}
\label{alg:imitating_desirable_behaviour}
\begin{algorithmic}[1]
\STATE Inputs: base policy $\policy$, environment $\env$, feedback model $\feedbackmodel$
\STATE Imitation dataset $\imitationdataset \gets$ behaviour cloning dataset
\FOR{$i = 1 \ldots N$}
    \STATE $\instruction \gets \mathsc{SampleInstruction}$
    \STATE $\traj_i \gets \rollout(\policy, \env, \instruction)$
    \FOR{verbalized observation $\verbalized_k$, action $\action_k$ in each step of $\traj_i$}
        \STATE $y_k = \feedbackmodel(\verbalized_k)$
        \IF{$y_k$ is desirable}
            \STATE $\imitationdataset \gets \imitationdataset \bigcup (\verbalized_k, \action_k)$
        \ENDIF
    \ENDFOR
\ENDFOR

\STATE Improved policy $\policy^\prime \gets \mathsc{TrainLM}(\imitationdataset)$
\end{algorithmic}
\end{algorithm}

\begin{algorithm}
\caption{Policy improvement using~\methodname s.}
\label{alg:improve_policy}
\begin{algorithmic}[1]
\STATE Inputs: base policy $\policy$, environment $\env$
\STATE Feedback model $\feedbackmodel \gets \mathsc{TrainFeedbackModel}(\policy, \llmmethod, \env)$
\STATE $\pi_0 \gets \pi$
\FOR{$k = 1 \ldots N$}
    \STATE $\pi_k \gets \mathsc{ImitateUsingFeedback}(\policy_{k-1}, \env, \feedbackmodel)$
\ENDFOR
\end{algorithmic}
\end{algorithm}

\newpage
\section{Comparison to~\textsc{Dagger}}
\label{app:dagger}

\begin{table}[!t]
    \centering
    \caption{
    Task completion rate on evaluation benchmarks, including~\textsc{Dagger}.
    }
    \label{tab:dagger}
    \begin{tabularx}{\linewidth}{XXXX}
    \toprule
         & ALFWorld & ScienceWorld & Touchdown \\
    \midrule
    \methodBehaviourCloning    &  62.6&  45.8& 57.5\\
    \methodActionPrediction    &  56.0& 39.0& 58.0\\
    \textsc{Dagger}    &  55.2& 22.5& 50.2\\
    \methodOurs    &  \textbf{64.1} & \textbf{47.1} & \textbf{59.7} \\
    \midrule
    \methodOursAdapt    &  74.6& 49.3& 62.8\\
    \bottomrule
    \end{tabularx}
\end{table}

Our main experiments in Section~\ref{sec:results} illustrate the difficulty of using LLMs as an expert to predict actions.
Specifically, we show that when these predictions are used for imitation learning, the resulting policy improvement is worse than using~\methodname s.
This performance degradation is exacerbated in environments with larger action spaces, such as ScienceWorld.

\textsc{Dagger}~\citep{rossReductionImitationLearning2011}~is an intermediate method between~\methodname s and using LLMs as an expert policies for imitation learning.
Specifically, in~\textsc{Dagger}, we also use LLMs as experts to predict action.
However, instead of using LLMs during each step, in~\textsc{Dagger} we use LLMs to provide batched retroactive action prediction similar to how in~\methodname s we use LLMs to provide batched retroactive feedback.
Here, we apply~\textsc{Dagger} action prediction to the exact same number of examples as when we collect feedback data for~\methodnameshort s.
In Table~\ref{tab:dagger}, we compare~\textsc{Dagger}~performance to those using LLM as an expert (\methodActionPrediction) and using~\methodname s (\methodOurs).
We find that although~\textsc{Dagger} is more efficient than~\methodActionPrediction~in that it annotates synthetic examples in batch, it underperforms~\methodActionPrediction~(and consequently~\methodOurs) across all benchmarks.

\newpage
\section{Quantitative and Qualitative Analyses of Learned Language Feedback}
\label{app:llama}

\begin{table}[!t]
    \centering
    \caption{Agreement between GPT-4 and~\textsc{Llama 2} across the benchmarks.
    We collect steps from rollouts on the training environments where either GPT-4 or~\textsc{Llama 2} identified a productive action.
    This table shows percentage of of those actions that are identified exclusively by GPT-4, exclusively by~\textsc{Llama 2}, and identified by both models.
    The total number of steps identfied are 40569 for ALFWorld, 68565 for ScienceWorld, and 90529 for Touchdown.
    }
    \label{tab:llama_confusion}
    \begin{tabularx}{\linewidth}{XXXX}
        \toprule
         & GPT-4 only & \textsc{Llama 2} only & both \\
         \midrule
         ALFWorld & 14.4\% & 49.3\% & 36.2\% \\ 
         ScienceWorld & 10.2\% & 62.3\% & 27.5\% \\ 
         Touchdown & 22.3\% & 67.3\% & 10.4\% \\ 
         \bottomrule
    \end{tabularx}
\end{table}

\begin{table}[!t]
    \centering
    \caption{Human verification of LLM feedback in terms of percentage of true positives and false positives.
    A true positive (TP) is a step that is correctly identified by the LLM as being productive to solving the task.
    A false positive (FP) is a step that is wrongly identified by the LLM as productive.
    We manually evaluate 10 examples from each benchmark, each with up to 20 steps.
    Support (\# of steps) is shown in brackets.
    }
    \label{tab:llm_manual}
    \begin{tabularx}{\linewidth}{XXX|XX}
        \toprule
         & \multicolumn{2}{c}{GPT-4} & \multicolumn{2}{c}{\textsc{Llama 2}} \\
         & TP & FP & TP & FP \\
         \midrule
         ALFWorld & 100\% (22) & 0 & 32\% (18) & 68\% (38) \\ 
         ScienceWorld & 78\% (38) & 22\% (11) & 48\% (38) & 52\% (41) \\ 
         Touchdown & 81\% (22) & 19\% (5) & 39\% (24) & 61\% (38) \\ 
         \bottomrule
    \end{tabularx}
\end{table}

\paragraph{Comparison of GPT-4 to~\textsc{Llama 2} 70B}
How much difference is there between language feedback obtained from the open-source~\textsc{Llama 2} vs from GPT-4?
Table~\ref{tab:llama_confusion}~shows that, surprisingly, there is a large degree of disagreement between GPT4 and~\textsc{Llama 2}.
Specifically,~\textsc{Llama 2} identifies significantly more actions as being productive to achieving the goal.

We perform a manual analysis of language feedback by GPT-4 and~\textsc{Llama 2} to characterize qualitative differences between feedback collected by these two models.
First, we roll out BC policies, then ask each model for feedback.
Each example contains a segment of up to 20 steps extracted from a rollout, and the LLM is prompted to list productive steps.
For each step the LLM identifies as productive to solving the task, we manually verify whether the step is indeed productive.
We manually inspect 10 examples from each model for each benchmark, for a total of $10 \times 2 \times 3 = 60$ examples.
Table~\ref{tab:llm_manual}~shows the number of true and false positives predicted by both models in this manual evaluation.
We find that a significant number of steps are incorrectly determined by~\textsc{Llama 2} as desirable.
When we train the policy on a combination of~\textsc{Llama 2} data and demonstrations used to learn the BC policy, we obtain worse task-completion percentage than using GPT-4 data and demonstrations.
Specially, performance drop from 64.1\% (GPT-4) to 56.0\% (\textsc{Llama 2}) on ALFWorld, 
from 47.1\% to 47.0\% on ScienceWorld,
and from 59.7\% to 56.5\% on Touchdown.

Table~\ref{tab:manual_feedback_examples}~shows some examples of steps identified as productive by these models that illustrate~\textsc{Llama 2}'s tendency to identify spurious actions as being productive.
In the ALFWorld examples, for instance,~\textsc{Llama 2} has a strong tendency to identify opening and closing cabinets and drawers as productive, even though they have nothing to do with putting a clean soap bar on the counter top (the first instruction) or putting a clean spatula on the side table (the second instruction).
Similarly in ScienceWorld,~\textsc{Llama 2} identifies unnecessary actions such as going outside (example 1) and going to the bedroom (example 2) as productive, even when the instruction explicitly details that the aluminum foil is found in the kitchen (example 1) and that the unknown substance is found in the workshop (example 2).
Finally,~\textsc{Llama 2} also tends to identify spurious actions as productive in Touchdown.
In the last example, the instruction asks to take a left after the first intersection, but~\textsc{Llama 2} rewards the left turn during the first turn, before the agent even arrives at the first intersection.
GPT-4, on the other hand, correctly identifies Step 8, when the agent finally encounters the first intersection, as productive.

We show in Section~\ref{sec:results} that small and cost-effective~\methodname s~are able to replicate LLM feedback through training.
Our comparison between GPT-4 and~\textsc{Llama 2} show that a less powerful model such as~\textsc{Llama 2} are unable to provide high-quality feedback.
The summary from this experiment are then that
1) powerful LLMs are necessary to provide good feedback, but expensive to run during on-line policy improvement
3) consequently, learning small~\methodnameshort s~is an effective solution to achieve high feedback performance while reducing inference cost during policy improvement.

\newpage
\section{GPU Usage}
\label{app:devices}
We train feedback models and policies using 80GB A100 GPUs.
To produce rollouts at in parallel, we use a cluster of 200 32GB V100 GPUs.
For all environments, feedback model training takes under 24 hours using one A100 GPU while inference can be performed locally using a 32GB GPU under 2 hours.
Policy training requires 1 day for ALFWorld, 2 days for ScienceWorld, and 3 days for Touchdown.
For all environments, policy rollout over the entire evaluation environments can be performed over the cluster of 200 32GB V100 GPUs in under 6 hours.

\newpage
\section{Correlation between LLM/\methodnameshort~feedback and true state values}
\label{sec:correlation}

To investigate whether LLM feedback is correlated with true state values, we obtained partial rollouts for environments in ALFWorld and asked GPT4 to score from 1-5 whether the partial rollouts are on the right track to solving the task.
We then ran a planner (with full observability) to complete these partial rollouts in order to obtain ground truth optimal values.
With no training, the LLM’s predicted score has strong correlation (0.61 Pearson) with the optimal values.
This means that GPT as a feedback model has strong zero-shot generalization on environments it was not trained on.

In addition, we regressed a FLAN-T5 model to estimate state values from language feedback using states from a random policy, then evaluated its predictions against true state values on mixed policies where an expert is select $p$ fraction of the time (and a random policy is used other times).
When we uniformly sample states from the random policy, for $p=0, 0.2, 0.4, 0.6, 0.8$, the regression model achieves Pearson correlations of $0.3-0.4$.
When we subsample training states evenly across values, we achieve Pearson $0.65 - 0.75$. This shows that state coverage, for which state value coverage is a proxy, is indeed important.
Conversely, when we use mixed policies for training as well, we obtain $0.78-0.85$ Pearson. This shows that optimality is indeed important.

\begin{table}[!th]
    \centering
    \small
    \caption{
    Example steps identified as productive by~\textcolor{darkblue}{GPT-4},\textcolor{darkred}{~\textsc{Llama 2}}, and~\textcolor{darkgreen}{both}.
    Touchdown steps are truncated for brevity.
    }
    \label{tab:manual_feedback_examples}
    \begin{tabularx}{\linewidth}{X}
        \toprule
        ALFWorld\\
        \midrule
        Task: clean some soapbar and put it in countertop.
        \\
        \textcolor{darkred}{
        Step 3
        Your action: take soapbottle 2 from toilet 1
        Result: You pick up the soapbottle 2 from the toilet 1.}
        \\
        \textcolor{darkgreen}{
        Step 6
        Your action: go to countertop 1
        Result: You arrive at loc 11. On the countertop 1, you see a candle 2...}
        \\
        \textcolor{darkred}{
        Step 11
        Your action: open cabinet 1
        Result: You open the cabinet 1. The cabinet 1 is open. In it, you...}
        \\
        \textcolor{darkred}{
        Step 17
        Your action: open cabinet 1
        Result: You open the cabinet 1. The cabinet 1 is open. In it, you see...}
        \\
        \textcolor{darkred}{
        Step 19
        Your action: go to cabinet 3
        Result: You arrive at loc 12. The cabinet 3 is closed.}
        \\
        Task: put a clean spatula in sidetable.
        \\
        \textcolor{darkred}{
        Step 3
        Your action: go to sidetable 1
        Result: You arrive at loc 22. On the sidetable 1, you see nothing.}
        \\
        \textcolor{darkred}{
        Step 10
        Your action: go to drawer 3
        Result: You arrive at loc 19. The drawer 3 is closed.}
        \\
        \textcolor{darkred}{
        Step 11
        Your action: open drawer 3
        Result: You open the drawer 3. The drawer 3 is open. In it, you...}
        \\
        \textcolor{darkred}{
        Step 14
        Your action: open drawer 4
        Result: You open the drawer 4. The drawer 4 is open. In it, you...}
        \\
        ...
        \\
        \midrule
        ScienceWorld \\
        \midrule
        Task: Your task is to determine if aluminum foil is electrically conductive. The aluminum foil is located around the kitchen. First, focus on the aluminum foil. If it is electrically conductive, place it in the yellow box. If it is electrically nonconductive, place it in the purple box. Do not focus on objects not central to completing the task!
        \\
        \textcolor{darkred}{
        Step 21
        Your action: open door to greenhouse
        Result: The door is now open...}
        \\
        \textcolor{darkred}{
        Step 23
        Your action: open door to outside
        Result: The door is now open...}
        \\
        \textcolor{darkred}{
        Step 25
        Your action: open door to foundry
        Result: The door is now open...}
        \\
        \textcolor{darkred}{
        Step 29
        Your action: go to outside
        Result: You move to the outside...}
        \\
        \textcolor{darkgreen}{
        Step 31
        Your action: go to kitchen
        Result: You move to the kitchen...}
        \\
        \textcolor{darkgreen}{
        Step 32
        Your action: pick up aluminum foil
        Result: You move the aluminum foil to the inventory...}
        \\
        \textcolor{darkred}{
        Step 33
        Your action: focus on aluminum foil
        Result: You focus on the aluminum foil...}
        \\
        \textcolor{darkblue}{
        Step 37
        Your action: go to workshop
        Result: You move to the workshop...}
        \\
        \textcolor{darkblue}{
        Step 38
        Your action: move aluminum foil to yellow box
        Result: You move the aluminum foil to the yellow...}
        \\
        Task: Your task is to determine if unknown substance N is electrically conductive. The unknown substance N is located around the workshop. First, focus on the unknown substance N. If it is electrically conductive, place it in the blue box. If it is electrically nonconductive, place it in the orange box. Do not focus on objects not central to completing the task!
        \\
        \textcolor{darkred}{
        Step 1
        Your action: open door to art studio
        Result: The door is now open...}
        \\
        \textcolor{darkred}{
        Step 4
        Your action: go to hallway
        Result: You move to the hallway...}
        \\
        \textcolor{darkred}{
        Step 6
        Your action: go to bedroom
        Result: You move to the bedroom...}
        \\
        \textcolor{darkred}{
        Step 9
        Your action: open door to hallway
        Result: The door is already open...}
        \\
        \textcolor{darkgreen}{
        Step 11
        Your action: open door to workshop
        Result: The door is now open...}
        \\
        \textcolor{darkblue}{
        Step 12
        Your action: go to workshop
        Result: You move to the workshop...}
        \\
        \textcolor{darkgreen}{
        Step 13
        Your action: pick up unknown substance
        Result: You move the unknown substance N to the...}
        \\
        \textcolor{darkgreen}{
        Step 15
        Your action: move unknown substance to blue box
        Result: You move the unknown substance N to...}
        \\
        \midrule
        Touchdown \\
        \midrule
        Task: Going with the flow of traffic, proceed to the intersection and take left. There should be some blue scaffolding on your right hand side, take a left here. You should now have a tall solid green fence on you right hand side, go straight to the next intersection and take a left. There's a bunch of scaffolding on the left hand side, go straight. Pass the first intersection you come across. You will pass mail trucks parked on both sides of the road.
        \\
        \textcolor{darkred}{
        Step 1
        Your action: slightly to your left 
        Result: behind you, you see:
parked, white box trucks
, two wider sidewalks
, this narrow two lane road
, the right sidewalk buildings
.
behind you, sightly to your left, you see:
, three air-conditioners
, three awning
, a smaller yellow taxi
.
to your left, you see:
, the theater awning
, a yellow cab car
, the second purple awning
.
slightly to your left, you see:
, a white-capped hydrant
, ornate gray balconies
, the purple wayfair truck
.
straight ahead, you see:
, a median strip
, some tall, brick buildings
, parked, white box trucks
, a blue bus lane sign
, the right sidewalk buildings
.
...}
\\
\textcolor{darkblue}{
Step 8
Your action: straight ahead
Result: behind you, you see:
, a brown storefront
, surface streets
, the right sidewalk buildings
, a parked black box truck
, some unremarkable brick buildings
.
behind you, sightly to your left, you see:
, then a storefront
, a white/grey van
, a large, blocky, gray building
.
to your left, you see:
, a large white store sign
, a construction vehicle
, a long gray and white 5 story building
slightly to your left, you see:
, some tall, brick buildings
, fedex van
, a white/grey van
.
straight ahead, you see:
, a large red brick apartment building
, an orange and white traffic object
, 3rd and 4th intersections
, a small blue car
, the right sidewalk buildings
, the parked yellow suv taxi
.
slightly to your right, you see:
...}
\\
        \bottomrule
    \end{tabularx}
\end{table}

\clearpage


\newpage
\section*{NeurIPS Paper Checklist}

\begin{enumerate}

\item {\bf Claims}
    \item[] Question: Do the main claims made in the abstract and introduction accurately reflect the paper's contributions and scope?
    \item[] Answer: \answerYes{} 
    \item[] Justification: we outline a new method for policy improvement via~\methodname. In the abstract and in the introduction, we claim that this method improves policy performance. Our experiments show that this method improves base policy performance on 3 distinct benchmarks. Furthermore, our experiments show that this technique outperforms alternative methods such as using LLMs as zeroshot policy, using LLMs to provide action labels, and using LLMs to provide~\textsc{Dagger}-style annotations. 

\item {\bf Limitations}
    \item[] Question: Does the paper discuss the limitations of the work performed by the authors?
    \item[] Answer: \answerYes{} 
    \item[] Justification: We discuss the limitations of this work in Appendix~\ref{app:limitations}.

\item {\bf Theory Assumptions and Proofs}
    \item[] Question: For each theoretical result, does the paper provide the full set of assumptions and a complete (and correct) proof?
    \item[] Answer: \answerNA{} 
    \item[] Justification: This work does not include theoretical results.

    \item {\bf Experimental Result Reproducibility}
    \item[] Question: Does the paper fully disclose all the information needed to reproduce the main experimental results of the paper to the extent that it affects the main claims and/or conclusions of the paper (regardless of whether the code and data are provided or not)?
    \item[] Answer: \answerYes{} 
    \item[] Justification: We outline experimental details in this paper, including steps and prompts required to produce LLM feedback to train~\methodnameshort. Furthermore, we outline hyperparameters and compute resources used to train feedback models and policies. Finally, we release our code and checkpoints for reproducing experimental results.

\item {\bf Open access to data and code}
    \item[] Question: Does the paper provide open access to the data and code, with sufficient instructions to faithfully reproduce the main experimental results, as described in supplemental material?
    \item[] Answer: \answerYes{} 
    \item[] Justification: We open source data and code required to reproduce our experiments. This release also contains documentation on how to run the code to reproduce data and re-train models described in this work. At this time, the release is anonymized at the link described in the introduction.

\item {\bf Experimental Setting/Details}
    \item[] Question: Does the paper specify all the training and test details (e.g., data splits, hyperparameters, how they were chosen, type of optimizer, etc.) necessary to understand the results?
    \item[] Answer: \answerYes{} 
    \item[] Justification: Our experiment section describe hyperparameters used to train models and collection feedback data.
    They are also contained in the source code release.

\item {\bf Experiment Statistical Significance}
    \item[] Question: Does the paper report error bars suitably and correctly defined or other appropriate information about the statistical significance of the experiments?
    \item[] Answer: \answerYes{} 
    \item[] Justification: We provide standard deviation of task completion rates in our main results table in Table~\ref{tab:main_results} across 3 random seeds.

\item {\bf Experiments Compute Resources}
    \item[] Question: For each experiment, does the paper provide sufficient information on the computer resources (type of compute workers, memory, time of execution) needed to reproduce the experiments?
    \item[] Answer: \answerYes{} 
    \item[] Justification: We describe compute usage in Appendix~\ref{app:devices}.
    
\item {\bf Code Of Ethics}
    \item[] Question: Does the research conducted in the paper conform, in every respect, with the NeurIPS Code of Ethics \url{https://neurips.cc/public/EthicsGuidelines}?
    \item[] Answer: \answerYes{} 
    \item[] Justification: we have reviewed the NeurIPS Code of Ethics and made sure that this work does not deviate from the Code of Ethics.

\item {\bf Broader Impacts}
    \item[] Question: Does the paper discuss both potential positive societal impacts and negative societal impacts of the work performed?
    \item[] Answer: \answerYes{} 
    \item[] Justification: We discuss the broader impact of this work in Appendix~\ref{app:impact}.
    
\item {\bf Safeguards}
    \item[] Question: Does the paper describe safeguards that have been put in place for responsible release of data or models that have a high risk for misuse (e.g., pretrained language models, image generators, or scraped datasets)?
    \item[] Answer: \answerNA{} 
    \item[] Justification: This work does not release data or models that have a high risk of misuse. We discuss limitations and impact of released models in Appendix~\ref{app:limitations} and Appendix~\ref{app:impact}.

\item {\bf Licenses for existing assets}
    \item[] Question: Are the creators or original owners of assets (e.g., code, data, models), used in the paper, properly credited and are the license and terms of use explicitly mentioned and properly respected?
    \item[] Answer: \answerYes{} 
    \item[] Justification: We detail license and terms of usage of models released for this work as well as assets used to develop this work in Appendix~\ref{app:licenses}.

\item {\bf New Assets}
    \item[] Question: Are new assets introduced in the paper well documented and is the documentation provided alongside the assets?
    \item[] Answer: \answerNA{} 
    \item[] Justification: This work does not release new assets.

\item {\bf Crowdsourcing and Research with Human Subjects}
    \item[] Question: For crowdsourcing experiments and research with human subjects, does the paper include the full text of instructions given to participants and screenshots, if applicable, as well as details about compensation (if any)? 
    \item[] Answer: \answerNA{} 
    \item[] Justification: This work does not involve crowdsourcing nor research with human subjects.

\item {\bf Institutional Review Board (IRB) Approvals or Equivalent for Research with Human Subjects}
    \item[] Question: Does the paper describe potential risks incurred by study participants, whether such risks were disclosed to the subjects, and whether Institutional Review Board (IRB) approvals (or an equivalent approval/review based on the requirements of your country or institution) were obtained?
    \item[] Answer: \answerNA{} 
    \item[] Justification: This work does not involve crowdsourcing nor research with human subjects.

\end{enumerate}

\end{document}